%% file: eccv26.tex
\documentclass{article}

\usepackage{eccv26}
\usepackage[utf8]{inputenc} 
\usepackage[T1]{fontenc}    
\usepackage{hyperref}       
\usepackage{url}            
\usepackage{booktabs}       
\usepackage{amsfonts}       
\usepackage{nicefrac}       
\usepackage{microtype}      
\usepackage{lipsum}
\usepackage{amsmath}
\usepackage{algorithm}
\usepackage{fancyhdr}       
\usepackage{graphicx}       
\graphicspath{{media/}}     
\usepackage{graphicx}
\usepackage{caption}
\usepackage{subcaption}
\usepackage{listings}
\usepackage{multirow}
\usepackage{algorithmic}

\newcommand{\etal}{\textit{et al}.}

\title{Input-Adaptive Generative Dynamics in Diffusion Models
}

\author{
    Yucheng Xing \\
    Department of Electrical and Computer Engineering \\
    Stony Brook University \\
    Stony Brook, NY 11794, USA \\
    \texttt{yucheng.xing@stonybrook.edu} \\
    \And
    Xiaodong Liu \\
    Department of Electrical and Computer Engineering \\
    Stony Brook University \\
    Stony Brook, NY 11794, USA \\
    \texttt{xiaodong.liu@stonybrook.edu} \\
    \And
    Xin Wang \\
    Department of Electrical and Computer Engineering \\
    Stony Brook University \\
    Stony Brook, NY 11794, USA \\
    \texttt{x.wang@stonybrook.edu} \\
}

\begin{document}
\maketitle


\input{Tex/abstract}


\input{Tex/sec1}

\input{Tex/sec2}

\input{Tex/sec3}

\input{Tex/sec4}

\input{Tex/sec5}

\bibliographystyle{splncs04}  
\bibliography{eccv26}

\end{document}

%% file: Tex/abstract.tex

\begin{abstract}


Diffusion models typically generate data through a fixed denoising trajectory that is shared across all samples. However, generation targets can differ in complexity, suggesting that a single pre-defined diffusion process may not be optimal for every input. In this work, we investigate \textit{input-adaptive generative dynamics} for diffusion models, where the generation process itself adapts to the conditions of each sample. Instead of relying on a fixed diffusion trajectory, the proposed framework allows the generative dynamics to adjust across inputs according to their generation requirements. To enable this behavior, we train the diffusion backbone under varying horizons and noise schedules, so that it can operate consistently under different input-adaptive trajectories. Experiments on conditional image generation show that diffusion trajectories can vary across inputs while maintaining generation quality and reducing the average number of sampling steps. These results provide a proof of the concept that diffusion processes can benefit from input-adaptive generative dynamics rather than relying on a single fixed trajectory.

\end{abstract}

%% file: Tex/sec1.tex

\section{Introduction~\label{sec:1}}


Generative modeling aims to learn mechanisms that can synthesize realistic data under complex conditions. Among recent advances, diffusion models have emerged as a powerful framework for high-quality image generation due to their stable training process and strong generative capability~\cite{ho2020denoising}. In diffusion models, data generation is formulated as a stochastic denoising process that gradually transforms noise into structured samples through a sequence of intermediate states. This formulation has enabled diffusion models to achieve strong performance across a wide range of generative tasks.

However, in most existing diffusion frameworks, the denoising trajectory is pre-defined and remains unchanged during generation. As a result, the same sequence of stochastic transformations is applied to all samples, regardless of their generation requirements. In practice, generation targets can differ in their structural complexity and semantic requirements. Some images may require longer or more detailed generative trajectories, while others can be synthesized with fewer refinement steps. This mismatch suggests that a single pre-defined diffusion trajectory may not always be optimal for every input.

This observation raises a natural question: \textit{can the generative dynamics of diffusion models adapt to the requirements of individual inputs?} In this work, we investigate the concept of \textit{\textbf{input-adaptive generative dynamics}}, where the diffusion process itself can adjust according to the conditions of each generation task. Instead of relying on a single shared generative trajectory, the proposed framework allows the diffusion dynamics to vary across inputs.

To investigate this idea, we develop a diffusion framework, \textit{\textbf{Adaptively Controllable Diffusion (AC-Diff)}}. The framework introduces mechanisms that estimate the required diffusion horizon for each sample and adjust the corresponding diffusion schedule accordingly. In addition, the model is trained with an adaptive sampling strategy that exposes the network to varying diffusion trajectories, enabling the generation process to adapt across inputs during inference.

Experiments on conditional image generation demonstrate that diffusion trajectories can vary across inputs while maintaining generation quality and reducing the average number of sampling steps. These results provide empirical evidence that diffusion models can benefit from input-adaptive generative dynamics rather than relying on a single fixed trajectory for all samples. 

The contributions of this work can be summarized as follows:
\begin{itemize}
    \item We introduce the concept of input-adaptive generative dynamics for diffusion models, where the generative trajectory can adapt to the requirements of individual inputs rather than remaining fixed.
    \item We develop a diffusion framework, AC-Diff, that enables sample-wise adaptation of the diffusion horizon and noise scheduling strategy. 
    \item Through experiments on conditional image generation, we demonstrate the feasibility of adaptive diffusion trajectories and show that the generation process can adjust across inputs while maintaining generation quality. 
\end{itemize}

The remainder of this paper is organized as follows: In Sec.~\ref{sec:2}, we overview the literature works related to diffusion models. The details of our model are given in Sec.~\ref{sec:3} and its effectiveness is proved in Sec.~\ref{sec:4} through experiments. Finally, we summarize the whole work in Sec.~\ref{sec:5}.

%% file: Tex/sec2.tex

\section{Related Work~\label{sec:2}}


\subsection{Conditional Diffusion Model~\label{sec:2.2}}

Early diffusion models are primarily developed as unconditional generative models, where samples are generated by reversing a stochastic noising process learned from data~\cite{ho2020denoising, song2020generative}. To make the generation controllable, many studies introduce external conditions to guide the reverse diffusion process. These conditions commonly include category labels~\cite{dhariwal2021diffusion, ho2022classifierfree, you2023diffusion}, textual descriptions~\cite{nichol2022glide, li2022upainting, liu2022control, gu2022vector, saharia2022photorealistic, ramesh2022hierarchical, Avrahami2023}, and reference images or structural signals~\cite{liu2022control, voynov2022sketchguided, yu2023freedom, batzolis2021conditional, zhang2023adding, qin2023unicontrol}. Existing conditional diffusion approaches can generally be categorized into \textit{guidance-based methods} and \textit{architecture-based conditioning methods}. Guidance-based approaches modify the sampling trajectory using guidance signals derived from classifiers or conditional score estimates. For example, classifier guidance introduces a separately trained classifier to steer the reverse diffusion process~\cite{dhariwal2021diffusion}, while classifier-free guidance removes the need for an external classifier by jointly training conditional and unconditional score estimators~\cite{ho2022classifierfree}. Text-guided generation often employs large multimodal encoders such as CLIP~\cite{radford2021learning} to measure the similarity between intermediate diffusion outputs and textual descriptions~\cite{nichol2022glide, li2022upainting, liu2022control}. Similar ideas have also been applied to image-based conditions, where additional encoders incorporate sketches or structural cues into the generation process~\cite{voynov2022sketchguided, batzolis2021conditional}. Architecture-based conditioning methods instead integrate conditions directly into the diffusion network. A representative example is latent diffusion~\cite{rombach2022highresolution}, where the diffusion process operates in a latent feature space and conditions are injected through cross-attention mechanisms. Similar conditioning strategies are also used in many text-to-image diffusion systems~\cite{li2022upainting, gu2022vector, saharia2022photorealistic}. Other works concatenate conditional embeddings with latent diffusion features~\cite{ramesh2022hierarchical, Avrahami2023}. More recent approaches incorporate spatial control signals such as edges or structural maps~\cite{zhang2023adding, qin2023unicontrol}. Structured conditions including scene graphs and layout representations have also been explored to guide image generation~\cite{yang2022diffusionbased, zheng2023layoutdiffusion}. In addition, some studies introduce conditions into different stages of the diffusion process, including the forward diffusion stage~\cite{zhang2023shiftddpms}. While these approaches improve controllability of the generated results, the diffusion trajectory itself is typically shared across inputs. In other words, conditions mainly influence what is generated, while the overall generative process remains largely consistent across samples.

\subsection{Efficient Diffusion Sampling~\label{sec:2.3}}

Despite their strong generative performance, diffusion models are computationally expensive because generation requires many iterative denoising steps~\cite{ho2020denoising, song2020generative}. A large body of work therefore focuses on accelerating diffusion sampling. One line of research reduces the number of sampling steps during inference. For example, sub-sampling strategies select a subset of diffusion steps from the original schedule~\cite{nichol2021improved, watson2021learning}. Nichol and Dhariwal~\cite{nichol2021improved} reduce inference cost by evenly sub-sampling the original diffusion schedule with learned variance estimation, while Watson~\etal~\cite{watson2021learning} formulate step selection as a dynamic programming problem. Song~\etal~\cite{song2022denoising} further generalize diffusion models with non-Markovian sampling processes that allow flexible transitions between diffusion steps. Other methods estimate the noise level of intermediate samples to adapt the sampling schedule during generation~\cite{sanroman2021noise}. Another line of work accelerates diffusion models through improved numerical solvers. From the perspective of continuous stochastic dynamics, the reverse diffusion process corresponds to solving stochastic differential equations, and sampling can be performed through probability flow ordinary differential equations~\cite{song2021scorebased}. This formulation motivates the development of efficient numerical solvers, including higher-order diffusion solvers and improved sampling schedules~\cite{karras2022elucidating, lu2022dpmsolver, jolicoeurmartineau2021gotta, dockhorn2022scorebased, zhang2023gddim, zhang2023fast, liu2022pseudo}. 
In addition, distillation techniques have been proposed to train student models that approximate multiple diffusion steps with fewer evaluations~\cite{luhman2021knowledge, salimans2022progressive, meng2023distillation}. Recent works~\cite{song2023consistency, geng2024consistency} further explore training strategies that enable one-step or few-step generation from diffusion models. Although these methods significantly improve efficiency, they typically use a shared sampling strategy across inputs, limiting the ability to allocate computation based on instance-specific generation difficulty.

\subsection{Adaptive Diffusion Dynamics~\label{sec:2.4}}

More recently, several studies have begun exploring adaptive mechanisms in diffusion models. 
For example, some approaches dynamically adjust diffusion schedules or step sizes based on numerical error estimation during sampling~\cite{lu2022dpmsolver, jolicoeurmartineau2021gotta, tang2024adadiff}. These methods introduce forms of adaptivity into the diffusion process, but the adjustments are typically performed locally during sampling. Other works explore instance-adaptive diffusion strategies that allocate different numbers of denoising steps according to input complexity. For example, Zhang~\etal propose AdaDiff~\cite{zhang2025adadiff}, learning a reinforcement learning policy that dynamically selects the number of diffusion steps during inference. These results further suggest that generation complexity may vary across inputs and that adaptive diffusion processes can improve efficiency. In contrast, our work investigates \textit{input-adaptive generative dynamics}, where the effective diffusion trajectory used for generation can vary according to the conditions of each generation task. Rather than performing step-wise adjustments during sampling or learning an inference-time step allocation policy, the generative dynamics are determined for each sample before the diffusion process begins, enabling sample-wise adaptive diffusion trajectories.

%% file: Tex/sec3.tex

\section{Methodology~\label{sec:3}}


We study input-adaptive generative dynamics for diffusion models, where the diffusion trajectory can depend on the generation conditions rather than remaining identical across inputs. This section first presents the formulation of input-adaptive diffusion dynamics. We then describe how the adaptive trajectory is instantiated and how the model is trained and applied for generation.

\subsection{Input-Adaptive Generative Dynamics~\label{sec:3.1}}

In diffusion-based generation, the sampling process can be viewed as a stochastic trajectory that progressively transforms noise into data through a sequence of denoising steps. This trajectory is typically characterized by two aspects: the number of diffusion steps that define the length of the process, and the noise dynamics that govern how the state evolves along the trajectory. 

Most existing diffusion models assume a fixed trajectory shared across all inputs. In particular, the total number of denoising steps $T$ and the associated noise schedule $\{\beta_{t}\}_{t=1}^{T}$ are pre-defined and remain identical for every generation task. While this design simplifies training and sampling, it also constrains the generative process to follow the same trajectory for all inputs. In practice, however, different generation conditions can vary substantially in semantic and structural complexity, which may require different levels of refinement during denoising.

In this work, we relax this assumption and allow the diffusion trajectory itself to depend on the generation conditions. From this perspective, the generative process is no longer a fixed procedure but a condition-dependent stochastic trajectory. Specifically, let $\mathbf{c}$ denote the conditioning information that specifies the generation task. Such conditioning information may include text prompts, structural signals, or other task-specific inputs. We represent the diffusion trajectory conditioned on the generation inputs as
\begin{equation}
    \tau(\mathbf{c}) = \left(T_{\text{cond}}, \{\beta'_{t}\}_{t=1}^{T_{\text{cond}}}\right),~\label{eq:input_adaptive_trajectory}
\end{equation}
where $T_{\text{cond}}$ determines the effective number of denoising steps for the given conditions, and $\{\beta'_{t}\}_{t=1}^{T_{\text{cond}}}$ specifies the noise schedule governing the stochastic dynamics along the trajectory. The conditional diffusion horizon is defined as
\begin{equation}
    T_{\text{cond}} = \mathcal{F}_{T}(\mathbf{c}),~\label{eq:T_cond_concept}
\end{equation}
where $\mathcal{F}_{T}(\cdot)$ estimates the required diffusion length according to the generation requirements. The associated noise dynamics are defined as
\begin{equation}
    \{\beta'_{t}\}_{t=1}^{T_{\text{cond}}} = \mathcal{F}_{\beta}(\mathbf{c}),~\label{eq:beta_cond_concept}
\end{equation}
which produces the noise schedule governing the stochastic evolution of the diffusion process. 

\begin{figure*}[!htpb]
    \centering
    \includegraphics[width=\linewidth]{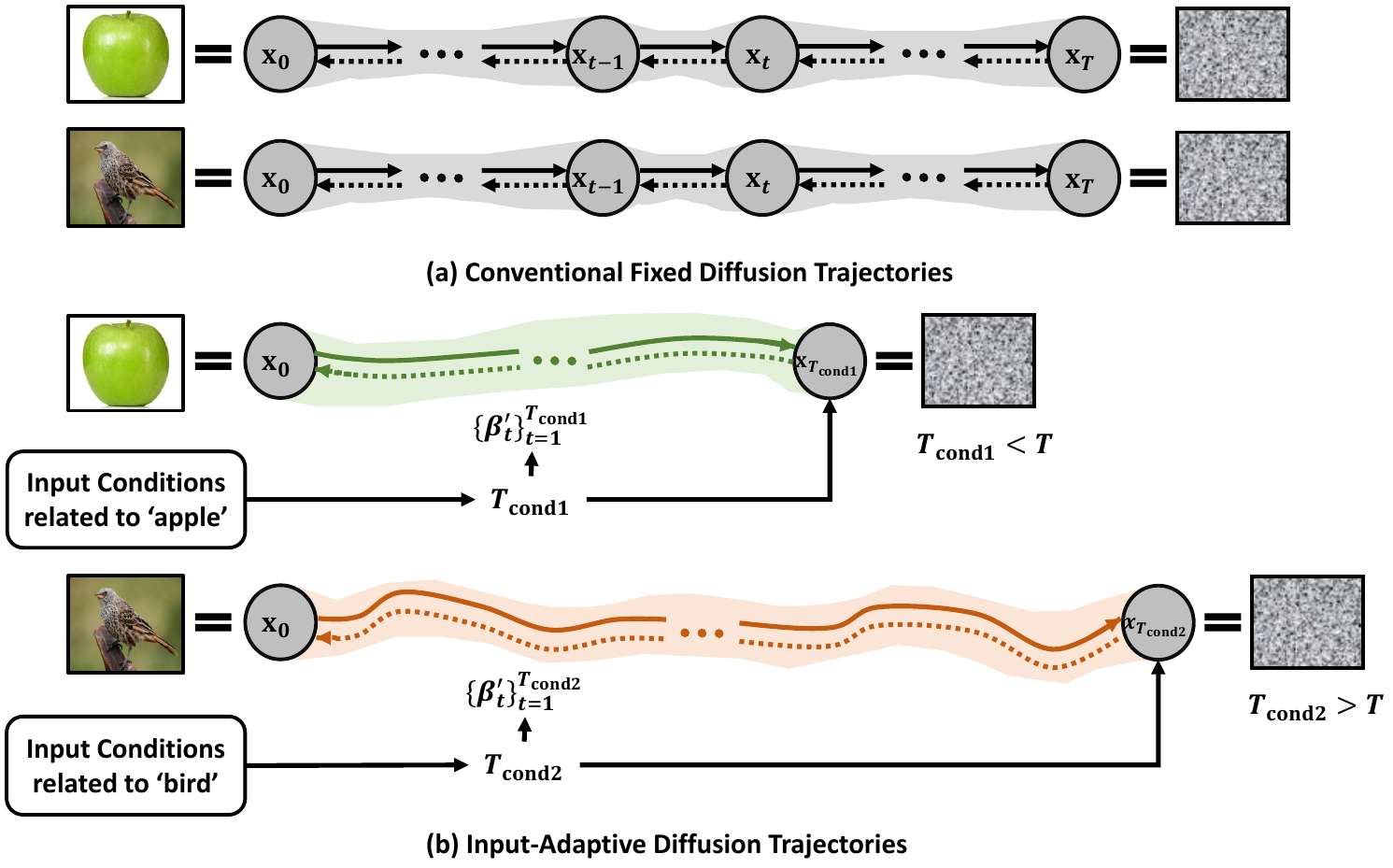}
    \caption{Illustration of diffusion trajectories. (a) Conventional diffusion models use a fixed trajectory shared across inputs. (b) Our method adopts input-adaptive diffusion trajectories.}
    \label{fig:adaptive_length}
\end{figure*}

Under this formulation, different inputs may follow diffusion trajectories with different lengths and stochastic dynamics, allowing the denoising process to adapt according to the generation conditions. Fig.~\ref{fig:adaptive_length} illustrates the difference between the fixed diffusion trajectories used in conventional diffusion models and the input-adaptive trajectories considered in our formulation.

\subsection{Conditional Diffusion Horizon Estimation~\label{sec:3.2}}

To instantiate the conditional diffusion horizon $T_{\text{cond}}$, we estimate the required diffusion length from the generation conditions. Intuitively, generation tasks involving richer structures or more detailed semantics may require longer denoising trajectories, while simpler tasks can often be synthesized with fewer refinement steps.

\begin{figure*}[!htpb]
    \centering
    \includegraphics[width=0.6\linewidth]{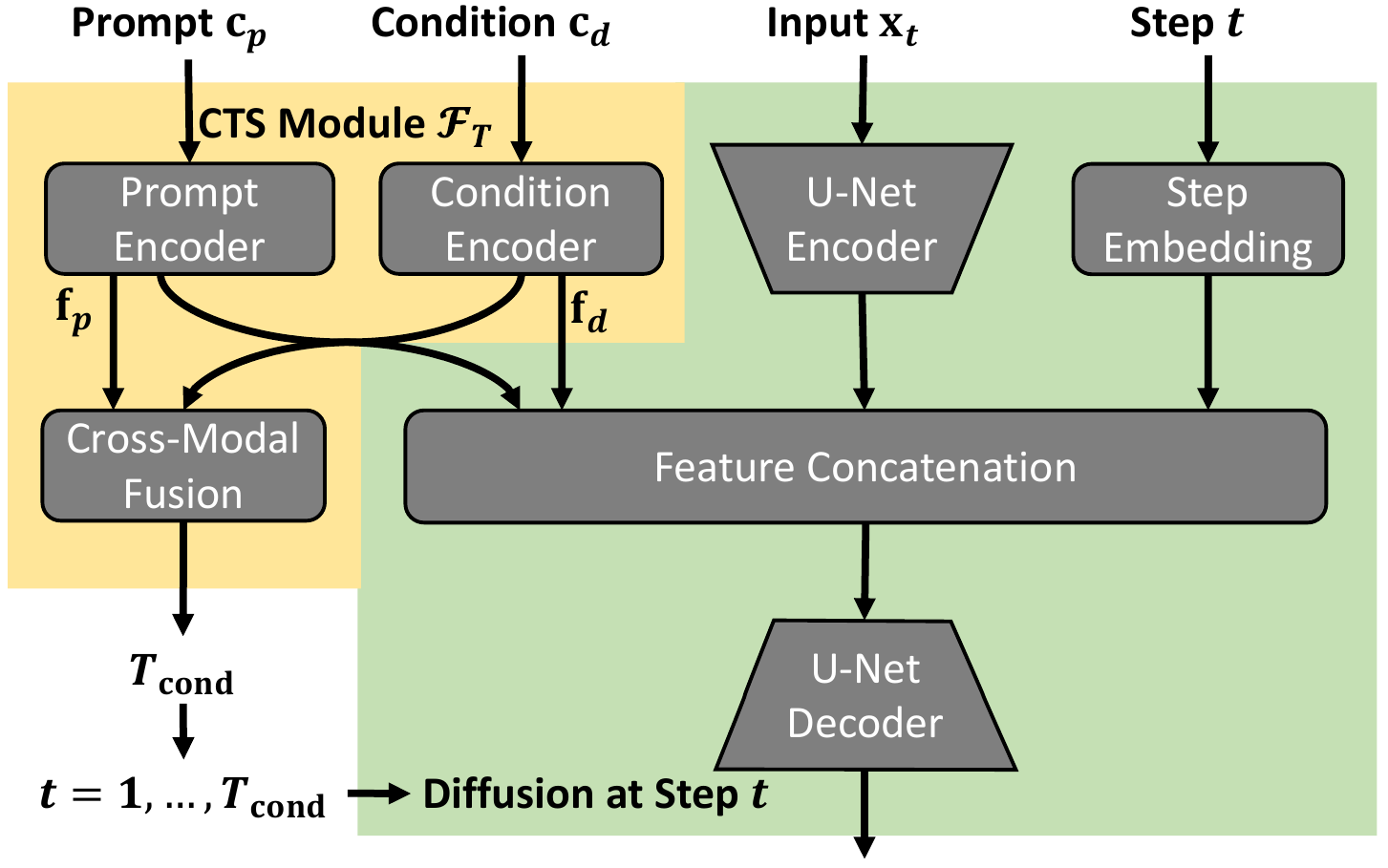}
    \caption{Structure of the CTS module for estimating the conditional diffusion horizon $T_{\text{cond}}$ from the generation conditions $(\mathbf{c}_p,\mathbf{c}_d)$.}
    \label{fig:cts}
\end{figure*}

We therefore introduce a conditional horizon estimator $\mathcal{F}_T(\cdot)$ that predicts the diffusion length from the conditioning information $\mathbf{c}$ as defined in Eq.~\ref{eq:T_cond_concept}. In practice, the conditioning information $\mathbf{c}$ may contain multiple modalities describing the generation task. In our implementation, $\mathbf{c}$ consists of two components: a text prompt $\mathbf{c}_p$ describing the semantic content and an additional structural condition $\mathbf{c}_d$ providing spatial guidance. The conditional diffusion horizon is therefore estimated as
\begin{equation}
    T_{\text{cond}}(\mathbf{c}_{p}, \mathbf{c}_{d}) = \mathcal{F}_{T}(\mathbf{c}_{p}, \mathbf{c}_{d}).~\label{eq:T_cond}
\end{equation}
To realize this estimator $\mathcal{F}_{T}(\cdot)$, we introduce a \textit{Conditional Time-Step (CTS) Module} that jointly analyzes the prompt and the structural condition, as illustrated in Fig.~\ref{fig:cts}. The CTS module extracts representations from both modalities and estimates the corresponding diffusion horizon.

Specifically, the text prompt $\mathbf{c}_{p}$ is encoded by a language encoder $\mathcal{E}_{p}(\cdot)$ to obtain the prompt embedding 
\begin{equation}
    \mathbf{f}_{p} = \mathcal{E}_{p}(\mathbf{c}_{p}),~\label{eq:prompt_enc}
\end{equation}
where $\mathcal{E}_{p}(\cdot)$ corresponds to the text transformer of a pre-trained CLIP model~\cite{radford2021learning}. Similarly, the structural condition $\mathbf{c}_{d}$ is encoded using a visual encoder $\mathcal{E}_{d}(\cdot)$ to produce the condition embedding
\begin{equation}
    \mathbf{f}_{d} = \mathcal{E}_{d}(\mathbf{c}_{d}),~\label{eq:condition_enc}
\end{equation}
where $\mathcal{E}_{d}(\cdot)$ is implemented using the visual transformer (ViT)~\cite{dosovitskiy2021image} component of the same CLIP model. Since CLIP is trained on paired image-text data, the prompt embedding $\mathbf{f}_{p}$ and the condition embedding $\mathbf{f}_{d}$ lie in a shared multimodal representation space. We concatenate the two embeddings and estimate the conditional diffusion horizon through a lightweight multi-layer perceptron $\mathcal{G}_{T}(\cdot)$ as
\begin{equation}
    T_{\text{cond}} = \mathcal{G}_{T}([\mathbf{f}_{p}, \mathbf{f}_{d}]).~\label{eq:crossmodel_fusion}
\end{equation}
Therefore, the conditional horizon estimator can be written as
\begin{equation}
    T_{\text{cond}} = \mathcal{F}_{T}(\mathbf{c}_{p}, \mathbf{c}_{d}) = \mathcal{G}_{T}([\mathcal{E}_{p}(\mathbf{c}_{p}), \mathcal{E}_{d}(\mathbf{c}_{d})]).~\label{eq:T_cond_full}
\end{equation}

In addition to semantic information, we further incorporate a spatial complexity measure derived from the structural condition. Given the conditional image $\mathbf{c}_{d}$, we compute an entropy-based spatial complexity ratio 
\begin{equation}
    r_s = -\sum_{v \in \mathbf{c}_d}p(v)\log p(v), ~\label{eq:r_s}
\end{equation}
where $v$ denotes pixel values and $p(v)$ represents their empirical distribution. Such entropy measures are widely used to characterize image complexity~\cite{larkin2016reflections, wu2013local, vila2014analysis}. The spatial complexity ratio $r_s$ is normalized (and optionally clipped) for numerical stability. The predicted diffusion horizon is then modulated as
\begin{equation}
    T_{\text{cond}} \leftarrow r_{s}\cdot T_{\text{cond}}.~\label{eq:final_T_cond}
\end{equation}
This conditional estimation of the diffusion horizon is performed only once for each generation task. Since the prompt and condition embeddings are also used by the diffusion model during generation, the additional computational overhead introduced by the CTS module is negligible.

\subsection{Adaptive Noise Dynamics~\label{sec:3.3}}

Adaptive noise dynamics correspond to the second component of the conditional diffusion trajectory introduced in Sec.~\ref{sec:3.1}. While Sec.~\ref{sec:3.2} estimates the conditional diffusion horizon $T_{\text{cond}}$, we now describe how the stochastic dynamics along the trajectory are determined. In our formulation, the noise schedule $\{\beta'_t\}_{t=1}^{T_{\text{cond}}}$ is produced by a condition-dependent mapping as described in Eq.~\ref{eq:beta_cond_concept}, which generates the diffusion noise dynamics according to the generation conditions.

\begin{figure}[!htpb]
    \centering
    \includegraphics[width=0.45\linewidth]{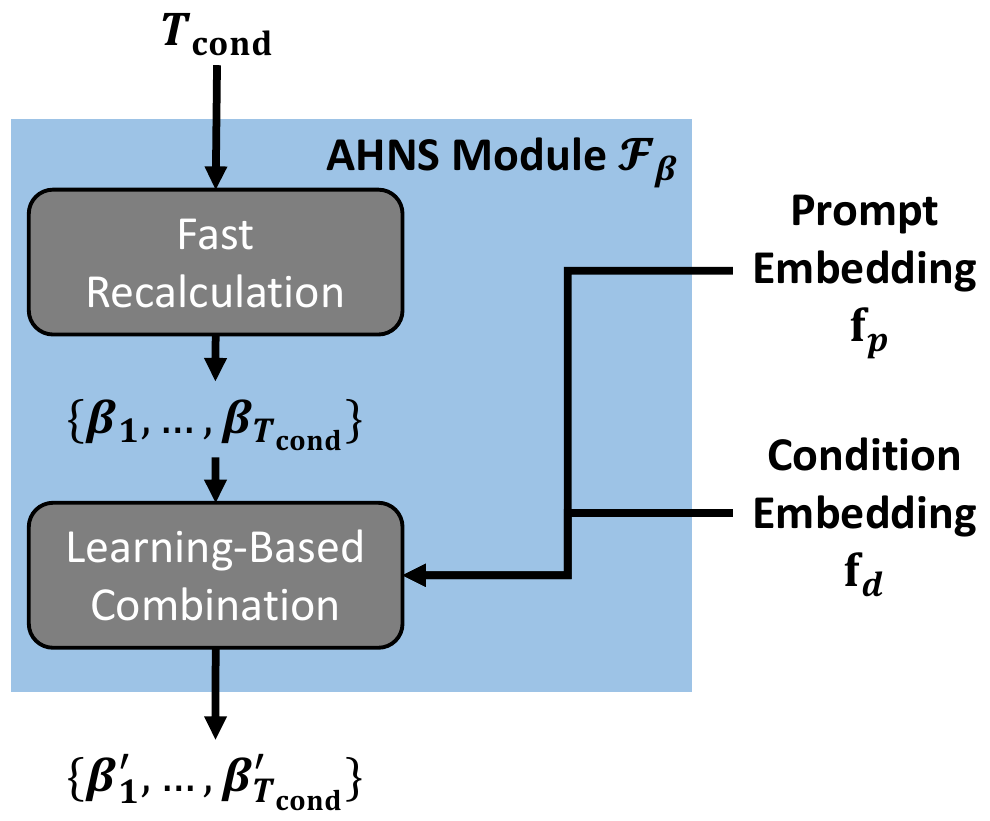}
    \caption{Structure of the AHNS module for generating the adaptive noise schedule $\{\beta'_t\}_{t=1}^{T_{\text{cond}}}$ conditioned on the predicted diffusion horizon and generation embeddings.}
    \label{fig:ahns}
\end{figure}

To instantiate this mapping, we introduce an \textit{Adaptive Hybrid Noise Scheduling (AHNS) Module} that constructs the conditional noise schedule in two stages, as illustrated in Fig.~\ref{fig:ahns}. The first stage determines a base noise schedule consistent with the estimated diffusion horizon, while the second stage further adapts the noise dynamics to the generation conditions.

\paragraph{\textbf{Fast Recalculation.}}
Given the conditional diffusion horizon $T_{\text{cond}}$, we first construct a base schedule $\{\beta_t\}_{t=1}^{T_{\text{cond}}}$ using a standard interpolation scheduler
\begin{equation}
    \{\beta_{t}\}_{t=1}^{T_{\text{cond}}} =\mathcal{S}(T_{\text{cond}}, \frac{\beta_{\text{min}}}{r_s},\frac{\beta_{\text{max}}}{r_s}),~\label{eq:beta_1}
\end{equation}
where $\mathcal{S}(\cdot)$ denotes a schedule generator such as linear, quadratic, or sigmoid interpolation. The boundary parameters $\beta_{\text{min}}$ and $\beta_{\text{max}}$ are scaled by the spatial complexity ratio $r_s$ introduced in Sec.~\ref{sec:3.2} (see Eq.~\ref{eq:r_s}), ensuring that the base schedule is consistent with both the diffusion horizon and the structural complexity of the inputs.

\paragraph{\textbf{Learning-Based Combination.}}
While the base schedule adapts to the trajectory length, additional flexibility can be introduced by allowing the reverse-process variance to vary with the generation conditions. Following~\cite{sohldickstein2015deep, ho2020denoising}, the parameters $\beta_t$ and $\tilde{\beta}_t = \frac{1-\overline{\alpha}_{t-1}}{1-\overline{\alpha}_t}\beta_t$ correspond to the upper and lower bounds of the reverse-process variance, where $\alpha_t=1-\beta_t$ and $\overline{\alpha}_t = \prod_{s=1}^{t}\alpha_s$. We therefore define the adaptive noise variance as a condition-dependent combination
\begin{equation}
    \sigma_t^2=\beta'_t=\lambda\beta_t+(1-\lambda)\tilde{\beta}_t,
\end{equation}
where the mixing coefficient $\lambda$ is predicted from the generation conditions using a lightweight neural predictor
\begin{equation}
    \lambda = \mathcal{G}_{\beta}([\mathbf{f}_{p},\mathbf{f}_{d}]),~\label{eq:lambda_g}
\end{equation}
with $\mathbf{f}_p$ and $\mathbf{f}_d$ denoting the prompt and condition embeddings introduced in Eq.~\ref{eq:prompt_enc} and Eq.~\ref{eq:condition_enc}. 

Combining the above components, the adaptive noise schedule can be written as
\begin{equation}
    \begin{aligned}
        \beta'_{t} &= \lambda\beta_{t} + (1-\lambda)\tilde{\beta}_{t} \\
                   &= \mathcal{G}_{\beta}([\mathbf{f}_{p},\mathbf{f}_{d}])\beta_{t} + (1-\mathcal{G}_{\beta}([\mathbf{f}_{p},\mathbf{f}_{d}]))\frac{1-\prod_{s=1}^{t-1}(1-\beta_{s})}{1-\prod_{s=1}^{t}(1-\beta_{s})}\beta_{t} ,~\label{eq:beta_2}
    \end{aligned}
\end{equation}
where $\beta_t\in\{\beta_{t}\}_{t=1}^{T_{\text{cond}}}$ are obtained from Eq.~\ref{eq:beta_1}.

\subsection{Framework Architecture~\label{sec:3.4}}

\begin{figure}[!htpb]
    \centering
    \includegraphics[width=0.7\linewidth]{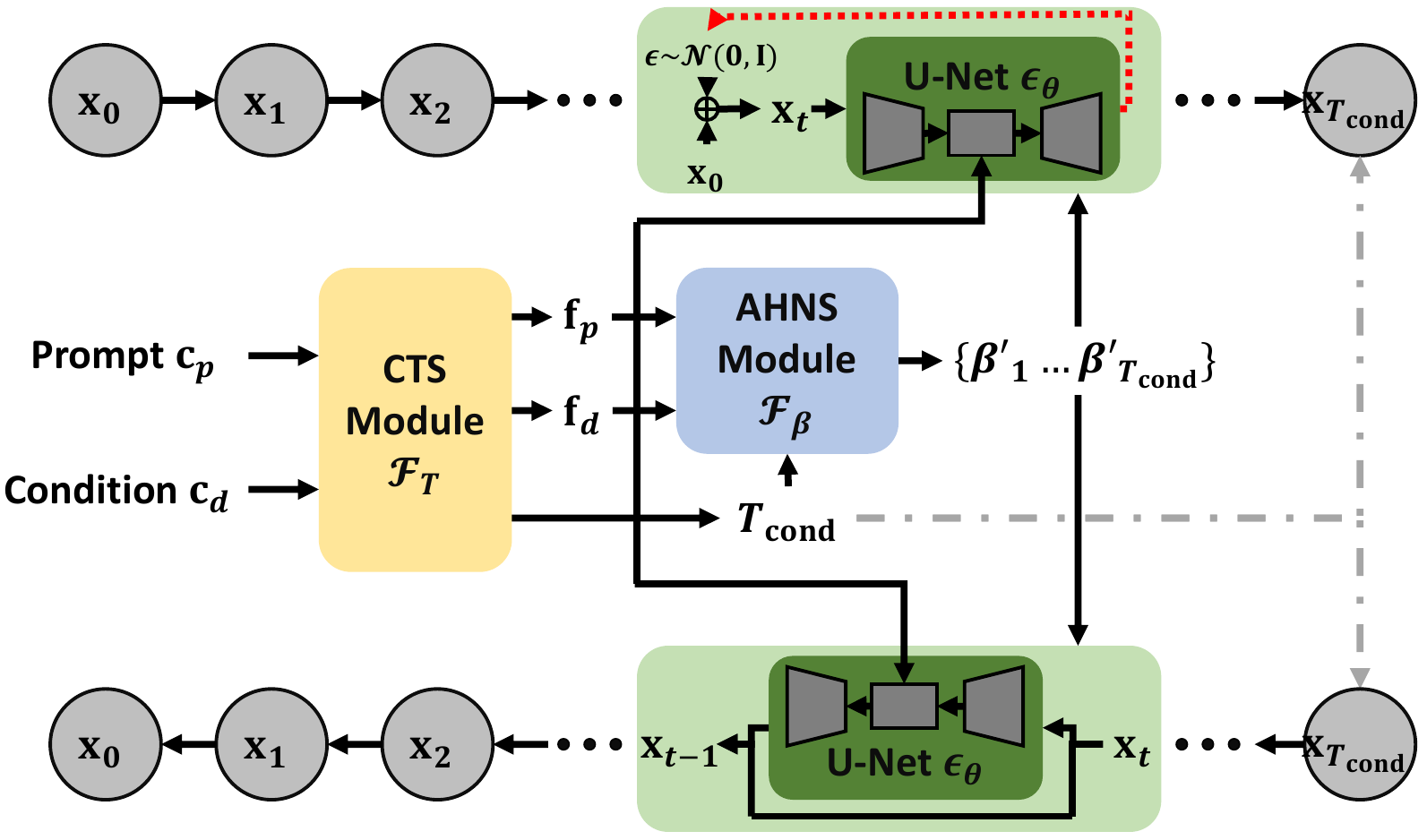}
    \caption{Overview of the proposed AC-Diff framework.}
    \label{fig:framework}
\end{figure}

To integrate the conditional diffusion horizon estimation and adaptive noise dynamics into diffusion generation, we construct the overall \textit{Adaptively Controllable Diffusion (AC-Diff)} framework illustrated in Fig.~\ref{fig:framework}. The framework consists of three main components: the Conditional Time-Step (CTS) module, the Adaptive Hybrid Noise Scheduling (AHNS) module, and a diffusion backbone network. 

Given the generation conditions, including the text prompt $\mathbf{c}_p$ and structural condition $\mathbf{c}_d$, the CTS module first extracts prompt and condition embeddings and predicts the conditional diffusion horizon $T_{\text{cond}}$ as described in Sec.~\ref{sec:3.2}. This estimated diffusion horizon determines the effective trajectory length for the diffusion process. Based on the predicted diffusion horizon and the conditioning embeddings, the AHNS module then constructs the adaptive noise schedule $\{\beta'_t\}_{t=1}^{T_{\text{cond}}}$ following the mechanism introduced in Sec.~\ref{sec:3.3}, which specifies the stochastic dynamics governing the diffusion trajectory.

Finally, the diffusion backbone, implemented using a conditional U-Net architecture, performs the denoising process using the adaptive trajectory parameters. During both training and generation, the U-Net receives the noisy sample $\mathbf{x}_t$, diffusion step $t$, and the conditioning embeddings as inputs, and predicts the noise component for the reverse diffusion process. The predicted noise is then used together with the adaptive noise schedule to update the diffusion state.

\subsection{Training and Generation~\label{sec:3.5}}

We now describe the training and generation procedures of the proposed framework. 
Training follows the standard diffusion objective, where the model learns to predict the noise added during the forward diffusion process. 
Given a clean image $\mathbf{x}_0$, the noisy sample $\mathbf{x}_t$ is obtained as
\begin{equation}
    \mathbf{x}_t = \sqrt{\overline{\alpha}'_t}\mathbf{x}_0 + \sqrt{1-\overline{\alpha}'_t}\mathbf{\epsilon}_t,~\label{eq:forward_process_2}
\end{equation}
where $\mathbf{\epsilon}_t \sim \mathcal{N}(\mathbf{0},\mathbf{I})$ and 
$\overline{\alpha}'_t = \prod_{s=1}^{t}(1-\beta'_s)$ is determined by the adaptive noise schedule introduced in Sec.~\ref{sec:3.3}. 
For each input, the adaptive schedule defines the coefficients of both the forward noising process and the reverse denoising process, ensuring that the diffusion trajectory used during generation is consistent with the one used during training.

Unlike conventional diffusion models that assume a fixed diffusion length $T$, our framework adopts the input-adaptive trajectory described in Sec.~\ref{sec:3.1}. 
Specifically, for each training sample with conditions $(\mathbf{c}_{p},\mathbf{c}_{d})$, the conditional diffusion horizon $T_{\text{cond}}$ and the corresponding noise schedule $\{\beta'_t\}_{t=1}^{T_{\text{cond}}}$ are first computed. 
The diffusion step $t$ is then sampled from the adaptive range $[1,T_{\text{cond}}]$, exposing the model to varying trajectory lengths during training and allowing it to learn generation dynamics that remain consistent with the adaptive trajectories used during inference. The resulting objective becomes
\begin{equation}
    \mathbb{E}_{t\in[1,T_{\text{cond}}(\mathbf{c}_p,\mathbf{c}_d)]}\left\|\mathbf{\epsilon}_t - \mathbf{\epsilon}_\theta(\mathbf{x}_t,t,\mathbf{c}_p,\mathbf{c}_d)\right\|^2,~\label{eq:objective_2}
\end{equation}
and the training procedure is summarized in~Algorithm.~\ref{alg:1}.
\begin{algorithm}[!htpb]
    \caption{Forward Diffusion Process}\label{alg:1}
    \begin{algorithmic}[1]
        \STATE \textbf{given} $\mathbf{x}_{0}, \mathbf{c}_{p}, \mathbf{c}_{d}$
        \REPEAT
        \STATE $T_{\text{cond}} \leftarrow Eq.~\ref{eq:T_cond_full}$
        \STATE $\{\beta'_{t}\}_{t=1}^{T_{\text{cond}}} \leftarrow Eq.~\ref{eq:beta_1}-Eq.~\ref{eq:beta_2}$
        \STATE $t \sim \{1, ..., T_{\text{cond}}\}$
        \STATE $\mathbf{\epsilon} \sim \mathcal{N}(\mathbf{0}, \mathbf{I})$
        \STATE Take gradient descent step on
        \begin{equation*}
            \nabla_{\theta} ||\mathbf{\epsilon} - \mathbf{\epsilon}_{\theta}(\sqrt{\overline{\alpha}'_{t}}\mathbf{x}_{0} + \sqrt{1 - \overline{\alpha}'_{t}}\mathbf{\epsilon},  t, \mathbf{c}_{p}, \mathbf{c}_{d})||^{2}
        \end{equation*}
        \UNTIL{converged}
    \end{algorithmic}
\end{algorithm}

During generation, the model first estimates the adaptive diffusion horizon $T_{\text{cond}}$ and constructs the corresponding noise schedule $\{\beta'_t\}_{t=1}^{T_{\text{cond}}}$ based on the given conditions $(\mathbf{c}_p,\mathbf{c}_d)$. 
Starting from Gaussian noise $\mathbf{x}_{T_{\text{cond}}}\sim\mathcal{N}(\mathbf{0},\mathbf{I})$, the reverse diffusion process iteratively samples
\begin{equation}
    \mathbf{x}_{t-1} = \frac{1}{\sqrt{\alpha'_t}}\left(\mathbf{x}_t - \frac{\beta'_t}{\sqrt{1 - \overline{\alpha}'_t}}\mathbf{\epsilon}_\theta(\mathbf{x}_t,t,\mathbf{c}_p,\mathbf{c}_d)\right) + \sqrt{\beta'_t}\mathbf{z},~\label{eq:backward_process}
\end{equation}p
where $\mathbf{z}\sim\mathcal{N}(\mathbf{0},\mathbf{I})$ when $t>1$ and $\mathbf{z}=\mathbf{0}$ otherwise. 
The complete generation procedure is summarized in~Algorithm.~\ref{alg:2}.

\begin{algorithm}[!htpb]
    \caption{Reverse Diffusion Process}\label{alg:2}
    \begin{algorithmic}[1]
        \STATE \textbf{given} $\mathbf{c}_{p}, \mathbf{c}_{d}$
        \STATE $T_{\text{cond}} \leftarrow Eq.~\ref{eq:T_cond_full}$
        \STATE $\{\beta'_{1}, ..., \beta'_{T_{\text{cond}}}\} \leftarrow Eq.~\ref{eq:beta_1}-Eq.~\ref{eq:beta_2}$
        \STATE $\mathbf{x}_{T_{\text{cond}}} \sim \mathcal{N}(\mathbf{0}, \mathbf{I})$
        \FOR{$t = T_{\text{cond}}, ..., 1$}
            \STATE $\mathbf{z} \sim \mathcal{N}(\mathbf{0}, \mathbf{I})$ \textbf{if} $t > 1$ \textbf{else} $\mathbf{z} = \mathbf{0}$
            \STATE $\mathbf{x}_{t - 1} \leftarrow Eq.~\ref{eq:backward_process}$
        \ENDFOR
        \STATE \textbf{return} $\mathbf{x}_{0}$
    \end{algorithmic}
\end{algorithm}

%% file: Tex/sec4.tex

\section{Experiment~\label{sec:4}}


\subsection{Experimental Setup~\label{sec:4.1}}

\subsubsection{Datasets}
CIFAR-10~\cite{cifar10} consists of $60,000$ color images from $10$ categories, each with a spatial resolution of $32\times32$. The dataset is divided into $50,000$ training images and $10,000$ testing images. In our experiments, the category name of each image is used as the text prompt, and the corresponding edge map is used as an additional structural condition to guide generation. All quantitative evaluations are conducted on the testing set.

\subsubsection{Evaluation Metrics}
We evaluate the proposed method from two aspects: generation quality and generation efficiency.

For generation quality, we report \textit{Fr\'echet Inception Distance (FID, $\downarrow$)}~\cite{heusel2018gans}, which measures the distributional similarity between generated and real images. 
To evaluate conditional alignment, we compute two CLIP-based scores~\cite{pmlr-v139-radford21a}: \textit{CS-t2i ($\uparrow$)}, which measures the similarity between generated images and text prompts, and \textit{CS-i2i ($\uparrow$)}, which evaluates the alignment between generated images and the input structural condition. We also report \textit{CLIP Aesthetic Score (C-Aes., $\uparrow$)}~\cite{schuhmann2022laionb} to reflect the perceptual quality of generated images.

For generation efficiency, we report \textit{Average Diffusion Time-Steps (Step, $\downarrow$)} and \textit{Average Execution Time (Time, $\downarrow$)}.

\subsubsection{Implementation Details}
In our experimental setting, the text prompt specifies the target category to be generated, while the input condition provides spatial guidance for the generation. Specifically, we use category names as text prompts and extract Canny edge maps from images as structural conditions, mimicking sketch-like guidance. To encode these inputs, we use the text and image encoders of a pre-trained CLIP ViT-B/32 model. The newly introduced components, including the cross-modal fusion module in CTS and the combination parameter predictor in AHNS, are implemented as two-layer MLPs with sigmoid activation. For fair comparison, all methods (including AC-Diff) are trained from scratch for $500\text{k}$ iterations with a batch size of $96$, and evaluated on a single NVIDIA RTX-4090 GPU.

\subsection{Overall Performance~\label{sec:4.2}}

\begin{table*}[!htpb]
    \centering
    \renewcommand\arraystretch{0.8}	
    \caption{Overall comparison among different methods on CIFAR-10.}
    \label{tab:overall_performance_cifar}
    \begin{tabular}{c c c c c c c}
        \toprule
            Method & FID($\downarrow$) & CS-t2i($\uparrow$) &CS-i2i ($\uparrow$) &C-Aes. ($\uparrow$) & Step($\downarrow$) & Time($\downarrow$) \\
            \midrule
            Original Image &- &0.2539 &0.7896 &3.7259 &- &- \\
            \midrule
            Unconditional Models \\
            \cmidrule(lr){1-1}
            DDPM~\cite{ho2020denoising} &29.5955 &0.2110 &0.7658 &3.5850 &1000 &15.1748 \\ 
            DDIM~\cite{song2022denoising} &29.6106 &0.2124 &0.7678 &3.6739 &1000 &16.4451 \\ 
                                          &30.9802 &0.2136 &0.7632 &3.6609 &100 &1.1036 \\
                                          &32.2251 &0.2160 &0.7633 &3.6106 &50 &0.5600 \\
            \midrule
            Conditional Models \\
            \cmidrule(lr){1-1}
            DDPM (cond r.) &32.1141 &0.2115 &0.7703 &3.5380 &1000 &12.9505 \\ 
            DDIM (cond r.) &34.6714 &0.2118 &0.7680 &3.6720 &1000 &16.1519 \\ 
                             &34.0696 &0.2125 &0.7752 &3.6640 &100 &1.3377 \\ 
                             &33.3599 &0.2140 &0.7670 &3.6298 &50 &0.7521 \\ 
            DDPM* (cond f.\&r.) &28.4607 &0.2546 &0.7955 &3.5899 &1000 &17.4065 \\ 
            DDIM* (cond f.\&r.) &28.6429 &0.2528 &0.7948 &3.7101 &1000 &18.4088 \\ 
                          &29.0791 &0.2556 &0.7913 &3.7087 &100 &1.9522 \\ 
                          &29.6803 &0.2575 &0.7898 &3.6722 &50 &0.9958 \\ 
            Guided-Diffusion (DDPM)~\cite{dhariwal2021diffusion} &42.4932 &0.2527 &0.7752 &3.1055 &250 &8.8107 \\ 
            Guided-Diffusion (DDIM)~\cite{dhariwal2021diffusion} &34.2655 &0.2348 &0.7719 &3.5243 &25 &0.8431 \\ 
            SDG~\cite{liu2022control} &30.2310 &0.2122 &0.7696 &3.4599 &1000 &55.0681 \\
            \midrule
            AC-Diff &22.4677 &0.2545 &0.7933 &3.7664 &141 &2.0376 \\
        \bottomrule
    \end{tabular}
\end{table*}

The overall comparison among different diffusion-based generative models is shown in Table.~\ref{tab:overall_performance_cifar}. Since our framework is built upon the DDPM formulation~\cite{ho2020denoising}, we first include the unconditional diffusion models DDPM~\cite{ho2020denoising} and DDIM~\cite{song2022denoising} as reference baselines. To evaluate conditional generation, we consider two strategies for incorporating the text prompts and image conditions. In the first strategy, the conditions are only injected during the reverse diffusion process while the model itself remains unconditionally trained, denoted as DDPM (cond r.) and DDIM (cond r.). In the second strategy, the conditions are incorporated during both training and generation, denoted as DDPM* (cond f.\&r.) and DDIM* (cond f.\&r.). In addition, we include representative conditional diffusion models, including Guided-Diffusion~\cite{dhariwal2021diffusion} and SDG~\cite{liu2022control}, for comparison. During evaluation, we generate $2.5$k images (approximately one quarter of the CIFAR-10 test set) and compute all evaluation metrics on the generated samples. It is worth noting that the reported FID values are obtained under a conditional generation setting with both text prompts and structural conditions. Moreover, the evaluation is conducted on $32\times32$ images using a limited number of generated samples, which can lead to absolute FID values that differ from those reported in unconditional CIFAR-10 generation benchmarks that typically use larger sample sets. Therefore, the reported FID values are not directly comparable with commonly reported unconditional CIFAR-10 results. As shown in Table.~\ref{tab:overall_performance_cifar}, AC-Diff achieves competitive generation quality while requiring fewer diffusion steps than conventional diffusion models. At the same time, the model maintains consistent conditional alignment with respect to both the text prompts and structural conditions, as reflected by the C-Score-t2i and C-Score-i2i metrics. Overall, these results suggest that the proposed adaptive diffusion trajectory provides an effective mechanism for improving generation efficiency while maintaining stable conditional generation performance.

\subsection{Ablation Study~\label{sec:4.3}}

We conduct ablation experiments to analyze the contributions of the key components in the proposed framework, including \textit{conditional training}, \textit{dynamic time-step}, and \textit{adaptive noise rescheduling}.

\noindent\textbf{Conditional Training.}
We first examine the role of incorporating conditional information during training. In Table.~\ref{tab:overall_performance_cifar}, DDPM and DDIM are compared with their variants that introduce conditional inputs only during generation, denoted as DDPM (cond r.) and DDIM (cond r.). These results indicate that directly injecting conditions into a pre-trained unconditional diffusion model during sampling provides limited improvements in conditional alignment and may lead to unstable generation quality. In contrast, when the prompt and structural conditions are incorporated during both training and generation, as in DDPM* (cond f.\&r.) and DDIM* (cond f.\&r.), the models are able to better utilize the conditional signals. As a result, both conditional alignment and visual quality become more stable. This observation suggests that learning conditional guidance during training helps the model effectively exploit the provided prompts and structural cues.

\begin{figure}[!htpb]
    \centering
    \includegraphics[width=0.7\linewidth]{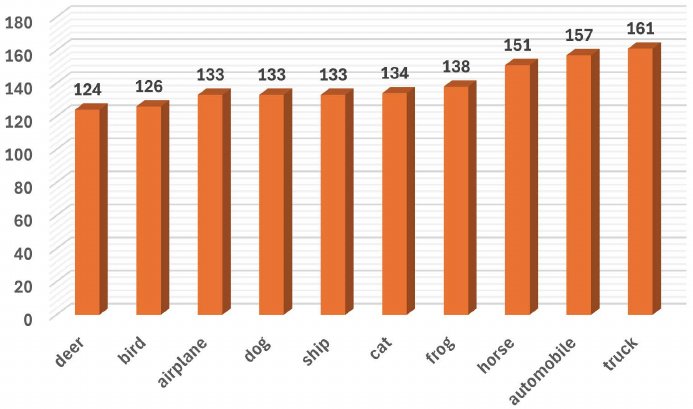}
    \caption{Average number of diffusion steps required for different CIFAR-10 categories under the adaptive trajectory.}
    \label{fig:category_level_step}
\end{figure}

\noindent\textbf{Dynamic Time-Step.}
The proposed AC-Diff introduces an adaptive mechanism to determine the diffusion trajectory length according to the input conditions. Experimental results show that images can be generated using substantially fewer diffusion steps while maintaining comparable generation quality. This observation suggests that a fixed large number of diffusion steps may not always be necessary for all generation tasks. To better illustrate this adaptive behavior, we report the category-level average number of diffusion steps required during generation in Fig.~\ref{fig:category_level_step}. Different image categories exhibit different diffusion horizons, indicating that the complexity of the generation task can vary across inputs. This observation supports the motivation of the proposed adaptive trajectory design, where the diffusion length is determined according to the input conditions rather than being fixed for all samples.

\noindent\textbf{Adaptive Noise Rescheduling.}
Since the proposed framework employs adaptive diffusion horizons, the corresponding noise schedule should also be adjusted accordingly. Intuitively, when fewer diffusion steps are used, each step needs to remove a larger portion of noise to maintain a consistent denoising trajectory. To evaluate this design, we compare two noise scheduling strategies: (1) an adaptive noise schedule that recalculates the noise ratios according to the estimated diffusion horizon, and (2) a fixed schedule obtained by directly downsampling the predefined diffusion schedule. The comparison results are shown in Table.~\ref{tab:adaptive_noise_schedule}. The adaptive noise scheduling strategy leads to more stable quality, indicating that adjusting the noise schedule according to the adaptive trajectory is beneficial.
\begin{table}[!htpb]
    \centering
    \renewcommand\arraystretch{0.8}
    \caption{Performance of AC-Diff with different noise rescheduling strategies.}
    \label{tab:adaptive_noise_schedule}
    \begin{tabular}{c c c c c}
        \toprule
            Method & FID ($\downarrow$) & CS-t2i ($\uparrow$) & CS-i2i ($\uparrow$) & C-Aes. ($\uparrow$) \\
        \midrule
            Fixed-$\beta$ &47.2681 &0.2499 &0.7927 &2.9297 \\
            Adaptive-$\beta$ &22.4677 &0.2545 &0.7933 &3.7664 \\
        \bottomrule
    \end{tabular}
\end{table}

\subsection{Qualitative Study~\label{sec:4.4}}

\begin{figure}[!htpb]
    \centering
    \includegraphics[width=0.8\linewidth]{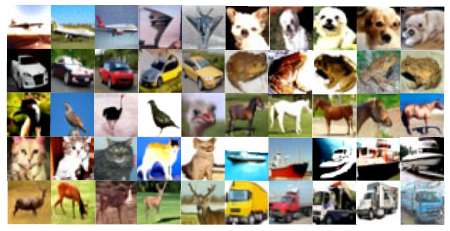}
    \caption{Examples of images generated by AC-Diff on CIFAR-10.}
    \label{fig:qualitative_res}
\end{figure}

We present representative examples of images generated by AC-Diff in Fig.~\ref{fig:qualitative_res}. The results demonstrate that the proposed framework is able to generate visually recognizable objects across different categories while preserving the structural cues provided by the input conditions. These qualitative results further illustrate that the proposed adaptive diffusion trajectory produces stable conditional generations across different categories.

%% file: Tex/sec5.tex

\section{Conclusion~\label{sec:5}}

In this paper, we propose an adaptive diffusion trajectory for conditional image generation, where the diffusion horizon is estimated according to the input prompts and structural conditions. By predicting the trajectory length and adaptively adjusting the noise schedule, the proposed approach avoids unnecessary diffusion steps while preserving generation quality for more complex samples. Experimental results on CIFAR-10 demonstrate that the proposed method achieves competitive generation performance while improving sampling efficiency compared with conventional diffusion models. In future work, we plan to extend the proposed approach to more complex datasets and broader conditional generation tasks.

%% file: eccv26.bib
@String(CVPR= {IEEE Conf. Comput. Vis. Pattern Recog.})

@String(ICCV= {Int. Conf. Comput. Vis.})

@String(ICIP = {IEEE Int. Conf. Image Process.})

@String(ICLR = {Int. Conf. Learn. Represent.})

@String(AAAI = {AAAI})

@String(CVPR  = {CVPR})

@String(ICCV  = {ICCV})

@String(ICIP  = {ICIP})

@String(ICLR  = {ICLR})

@inproceedings{ho2020denoising,
       author = {Ho, Jonathan and Jain, Ajay and Abbeel, Pieter},
 booktitle = {Advances in Neural Information Processing Systems},
 editor = {H. Larochelle and M. Ranzato and R. Hadsell and M.F. Balcan and H. Lin},
 pages = {6840--6851},
 publisher = {Curran Associates, Inc.},
 title = {Denoising Diffusion Probabilistic Models},
 url = {https://proceedings.neurips.cc/paper_files/paper/2020/file/4c5bcfec8584af0d967f1ab10179ca4b-Paper.pdf},
 volume = {33},
 year = {2020}
}

@article{song2023consistency,
  title={Consistency models},
  author={Song, Yang and Dhariwal, Prafulla and Chen, Mark and Sutskever, Ilya},
  year={2023}
}

@article{geng2024consistency,
  title={Consistency models made easy},
  author={Geng, Zhengyang and Pokle, Ashwini and Luo, William and Lin, Justin and Kolter, J Zico},
  journal={arXiv preprint arXiv:2406.14548},
  year={2024}
}

@inproceedings{tang2024adadiff,
  title={Adadiff: Accelerating diffusion models through step-wise adaptive computation},
  author={Tang, Shengkun and Wang, Yaqing and Ding, Caiwen and Liang, Yi and Li, Yao and Xu, Dongkuan},
  booktitle={European Conference on Computer Vision},
  pages={73--90},
  year={2024},
  organization={Springer}
}

@inproceedings{zhang2025adadiff,
  title={AdaDiff: adaptive step selection for fast diffusion models},
  author={Zhang, Hui and Wu, Zuxuan and Xing, Zhen and Shao, Jie and Jiang, Yu-Gang},
  booktitle={Proceedings of the AAAI Conference on Artificial Intelligence},
  volume={39},
  number={9},
  pages={9914--9922},
  year={2025}
}

@inproceedings{dhariwal2021diffusion,
title={Diffusion Models Beat {GAN}s on Image Synthesis},
author={Prafulla Dhariwal and Alexander Quinn Nichol},
booktitle={Advances in Neural Information Processing Systems},
editor={A. Beygelzimer and Y. Dauphin and P. Liang and J. Wortman Vaughan},
year={2021},
url={https://openreview.net/forum?id=AAWuCvzaVt}
}

@inproceedings{
ho2022classifierfree,
title={Classifier-Free Diffusion Guidance},
author={Jonathan Ho and Tim Salimans},
booktitle={NeurIPS 2021 Workshop on Deep Generative Models and Downstream Applications},
year={2021},
url={https://openreview.net/forum?id=qw8AKxfYbI}
}

@article{nichol2022glide,
    title={Glide: Towards photorealistic image generation and editing with text-guided diffusion models},
  author={Nichol, Alex and Dhariwal, Prafulla and Ramesh, Aditya and Shyam, Pranav and Mishkin, Pamela and McGrew, Bob and Sutskever, Ilya and Chen, Mark},
  journal={arXiv preprint arXiv:2112.10741},
  year={2021}
}

@article{li2022upainting,
      title={UPainting: Unified Text-to-Image Diffusion Generation with Cross-modal Guidance}, 
      author={Wei Li and Xue Xu and Xinyan Xiao and Jiachen Liu and Hu Yang and Guohao Li and Zhanpeng Wang and Zhifan Feng and Qiaoqiao She and Yajuan Lyu and Hua Wu},
      year={2022},
      journal={arXiv:2210.16031 [cs.CV]}
}

@InProceedings{liu2022control,
    author    = {Liu, Xihui and Park, Dong Huk and Azadi, Samaneh and Zhang, Gong and Chopikyan, Arman and Hu, Yuxiao and Shi, Humphrey and Rohrbach, Anna and Darrell, Trevor},
    title     = {More Control for Free! Image Synthesis With Semantic Diffusion Guidance},
    booktitle = {Proceedings of the IEEE/CVF Winter Conference on Applications of Computer Vision (WACV)},
    month     = {January},
    year      = {2023},
    pages     = {289-299}
}

@inproceedings{voynov2022sketchguided,
author = {Voynov, Andrey and Aberman, Kfir and Cohen-Or, Daniel},
title = {Sketch-Guided Text-to-Image Diffusion Models},
year = {2023},
isbn = {9798400701597},
publisher = {Association for Computing Machinery},
address = {New York, NY, USA},
url = {https://doi.org/10.1145/3588432.3591560},
doi = {10.1145/3588432.3591560},
booktitle = {ACM SIGGRAPH 2023 Conference Proceedings},
articleno = {55},
numpages = {11},
location = {Los Angeles, CA, USA},
series = {SIGGRAPH '23}
}

@article{yu2023freedom,
title={FreeDoM: Training-Free Energy-Guided Conditional Diffusion Model},
author={Yu, Jiwen and Wang, Yinhuai and Zhao, Chen and Ghanem, Bernard and Zhang, Jian},
journal={Proceedings of the IEEE/CVF International Conference on Computer Vision (ICCV)},
year={2023}
}

@article{batzolis2021conditional,
      title={Conditional Image Generation with Score-Based Diffusion Models}, 
      author={Georgios Batzolis and Jan Stanczuk and Carola-Bibiane Schönlieb and Christian Etmann},
      year={2021},
      journal={arXiv:2111.13606 [cs.LG]}
}

@InProceedings{rombach2022highresolution,
    author    = {Rombach, Robin and Blattmann, Andreas and Lorenz, Dominik and Esser, Patrick and Ommer, Bj\"orn},
    title     = {High-Resolution Image Synthesis With Latent Diffusion Models},
    booktitle = {Proceedings of the IEEE/CVF Conference on Computer Vision and Pattern Recognition (CVPR)},
    month     = {June},
    year      = {2022},
    pages     = {10684-10695}
}

@InProceedings{gu2022vector,
    author    = {Gu, Shuyang and Chen, Dong and Bao, Jianmin and Wen, Fang and Zhang, Bo and Chen, Dongdong and Yuan, Lu and Guo, Baining},
    title     = {Vector Quantized Diffusion Model for Text-to-Image Synthesis},
    booktitle = {Proceedings of the IEEE/CVF Conference on Computer Vision and Pattern Recognition (CVPR)},
    month     = {June},
    year      = {2022},
    pages     = {10696-10706}
}

@inproceedings{
saharia2022photorealistic,
title={Photorealistic Text-to-Image Diffusion Models with Deep Language Understanding},
author={Chitwan Saharia and William Chan and Saurabh Saxena and Lala Li and Jay Whang and Emily Denton and Seyed Kamyar Seyed Ghasemipour and Raphael Gontijo-Lopes and Burcu Karagol Ayan and Tim Salimans and Jonathan Ho and David J. Fleet and Mohammad Norouzi},
booktitle={Advances in Neural Information Processing Systems},
editor={Alice H. Oh and Alekh Agarwal and Danielle Belgrave and Kyunghyun Cho},
year={2022},
url={https://openreview.net/forum?id=08Yk-n5l2Al}
}

@article{ramesh2022hierarchical,
  title={Hierarchical Text-Conditional Image Generation with CLIP Latents},
  author={Aditya Ramesh and Prafulla Dhariwal and Alex Nichol and Casey Chu and Mark Chen},
  journal={ArXiv},
  year={2022},
  volume={abs/2204.06125},
  url={https://api.semanticscholar.org/CorpusID:248097655}
}

@inproceedings{Avrahami2023,
  title={Spatext: Spatio-textual representation for controllable image generation},
  author={Avrahami, Omri and Hayes, Thomas and Gafni, Oran and Gupta, Sonal and Taigman, Yaniv and Parikh, Devi and Lischinski, Dani and Fried, Ohad and Yin, Xi},
  booktitle={Proceedings of the IEEE/CVF Conference on Computer Vision and Pattern Recognition},
  pages={18370--18380},
  year={2023}
}

@inproceedings{zhang2023adding,
  title={Adding Conditional Control to Text-to-Image Diffusion Models}, 
  author={Lvmin Zhang and Anyi Rao and Maneesh Agrawala},
  booktitle={IEEE International Conference on Computer Vision (ICCV)},
  year={2023}
}

@article{qin2023unicontrol,
  title={UniControl: A Unified Diffusion Model for Controllable Visual Generation In the Wild},
  author={Qin, Can and Zhang, Shu and Yu, Ning and Feng, Yihao and Yang, Xinyi and Zhou, Yingbo and Wang, Huan and Niebles, Juan Carlos and Xiong, Caiming and Savarese, Silvio and others},
  journal={arXiv preprint arXiv:2305.11147},
  year={2023}
}

@article{yang2022diffusionbased,
  title={Diffusion-based scene graph to image generation with masked contrastive pre-training},
  author={Yang, Ling and Huang, Zhilin and Song, Yang and Hong, Shenda and Li, Guohao and Zhang, Wentao and Cui, Bin and Ghanem, Bernard and Yang, Ming-Hsuan},
  journal={arXiv preprint arXiv:2211.11138},
  year={2022}
}

@InProceedings{zheng2023layoutdiffusion,
    author    = {Zheng, Guangcong and Zhou, Xianpan and Li, Xuewei and Qi, Zhongang and Shan, Ying and Li, Xi},
    title     = {LayoutDiffusion: Controllable Diffusion Model for Layout-to-Image Generation},
    booktitle = {Proceedings of the IEEE/CVF Conference on Computer Vision and Pattern Recognition (CVPR)},
    month     = {June},
    year      = {2023},
    pages     = {22490-22499}
}

@article{you2023diffusion,
      title={Diffusion Models and Semi-Supervised Learners Benefit Mutually with Few Labels}, 
      author={Zebin You and Yong Zhong and Fan Bao and Jiacheng Sun and Chongxuan Li and Jun Zhu},
      year={2023},
      archivePrefix={arXiv:2302.10586 [cs.CV]}
}

@article{zhang2023shiftddpms,
  title={Shiftddpms: Exploring conditional diffusion models by shifting diffusion trajectories},
  author={Zhang, Zijian and Zhao, Zhou and Yu, Jun and Tian, Qi},
  journal={arXiv preprint arXiv:2302.02373},
  year={2023}
}

@inproceedings{song2020generative,
 author = {Song, Yang and Ermon, Stefano},
 booktitle = {Advances in Neural Information Processing Systems},
 editor = {H. Wallach and H. Larochelle and A. Beygelzimer and F. d\textquotesingle Alch\'{e}-Buc and E. Fox and R. Garnett},
 pages = {},
 publisher = {Curran Associates, Inc.},
 title = {Generative Modeling by Estimating Gradients of the Data Distribution},
 url = {https://proceedings.neurips.cc/paper_files/paper/2019/file/3001ef257407d5a371a96dcd947c7d93-Paper.pdf},
 volume = {32},
 year = {2019}
}

@InProceedings{nichol2021improved,
  title = 	 {Improved Denoising Diffusion Probabilistic Models},
  author =       {Nichol, Alexander Quinn and Dhariwal, Prafulla},
  booktitle = 	 {Proceedings of the 38th International Conference on Machine Learning},
  pages = 	 {8162--8171},
  year = 	 {2021},
  editor = 	 {Meila, Marina and Zhang, Tong},
  volume = 	 {139},
  series = 	 {Proceedings of Machine Learning Research},
  month = 	 {18--24 Jul},
  publisher =    {PMLR},
  url = 	 {https://proceedings.mlr.press/v139/nichol21a.html}
}

@misc{
watson2021learning,
title={Learning to Efficiently Sample from Diffusion Probabilistic Models},
author={Daniel Watson and Jonathan Ho and Mohammad Norouzi and William Chan},
year={2022},
url={https://openreview.net/forum?id=LOz0xDpw4Y}
}

@inproceedings{
song2022denoising,
title={Denoising Diffusion Implicit Models},
author={Jiaming Song and Chenlin Meng and Stefano Ermon},
booktitle={International Conference on Learning Representations},
year={2021},
url={https://openreview.net/forum?id=St1giarCHLP}
}

@article{sanroman2021noise,
      title={Noise Estimation for Generative Diffusion Models}, 
      author={Robin San-Roman and Eliya Nachmani and Lior Wolf},
      year={2021},
      archivePrefix={arXiv:2104.02600 [cs.LG]}
}

@inproceedings{
song2021scorebased,
title={Score-Based Generative Modeling through Stochastic Differential Equations},
author={Yang Song and Jascha Sohl-Dickstein and Diederik P Kingma and Abhishek Kumar and Stefano Ermon and Ben Poole},
booktitle={International Conference on Learning Representations},
year={2021},
url={https://openreview.net/forum?id=PxTIG12RRHS}
}

@inproceedings{
karras2022elucidating,
title={Elucidating the Design Space of Diffusion-Based Generative Models},
author={Tero Karras and Miika Aittala and Timo Aila and Samuli Laine},
booktitle={Advances in Neural Information Processing Systems},
editor={Alice H. Oh and Alekh Agarwal and Danielle Belgrave and Kyunghyun Cho},
year={2022},
url={https://openreview.net/forum?id=k7FuTOWMOc7}
}

@inproceedings{
zhang2023fast,
title={Fast Sampling of Diffusion Models with Exponential Integrator},
author={Qinsheng Zhang and Yongxin Chen},
booktitle={The Eleventh International Conference on Learning Representations },
year={2023},
url={https://openreview.net/forum?id=Loek7hfb46P}
}

@inproceedings{
lu2022dpmsolver,
title={{DPM}-Solver: A Fast {ODE} Solver for Diffusion Probabilistic Model Sampling in Around 10 Steps},
author={Cheng Lu and Yuhao Zhou and Fan Bao and Jianfei Chen and Chongxuan Li and Jun Zhu},
booktitle={Advances in Neural Information Processing Systems},
editor={Alice H. Oh and Alekh Agarwal and Danielle Belgrave and Kyunghyun Cho},
year={2022},
url={https://openreview.net/forum?id=2uAaGwlP_V}
}

@article{jolicoeurmartineau2021gotta,
      title={Gotta Go Fast When Generating Data with Score-Based Models}, 
      author={Alexia Jolicoeur-Martineau and Ke Li and R{\'e}mi Pich{\'e}-Taillefer and Tal Kachman and Ioannis Mitliagkas},
      journal={arXiv preprint arXiv:2105.14080},
      year={2021}
}

@inproceedings{
zhang2023gddim,
title={g{DDIM}: Generalized denoising diffusion implicit models},
author={Qinsheng Zhang and Molei Tao and Yongxin Chen},
booktitle={The Eleventh International Conference on Learning Representations },
year={2023},
url={https://openreview.net/forum?id=1hKE9qjvz-}
}

@inproceedings{
liu2022pseudo,
title={Pseudo Numerical Methods for Diffusion Models on Manifolds},
author={Luping Liu and Yi Ren and Zhijie Lin and Zhou Zhao},
booktitle={International Conference on Learning Representations},
year={2022},
url={https://openreview.net/forum?id=PlKWVd2yBkY}
}

@inproceedings{dockhorn2022scorebased,
    title={Score-Based Generative Modeling with Critically-Damped Langevin Diffusion},
    author={Tim Dockhorn and Arash Vahdat and Karsten Kreis},
    booktitle={International Conference on Learning Representations (ICLR)},
    year={2022}
}

@article{luhman2021knowledge,
      title={Knowledge Distillation in Iterative Generative Models for Improved Sampling Speed}, 
      author={Eric Luhman and Troy Luhman},
      year={2021},
      archivePrefix={arXiv:2101.02388 [cs.LG]}
}

@inproceedings{
salimans2022progressive,
title={Progressive Distillation for Fast Sampling of Diffusion Models},
author={Tim Salimans and Jonathan Ho},
booktitle={International Conference on Learning Representations},
year={2022},
url={https://openreview.net/forum?id=TIdIXIpzhoI}
}

@inproceedings{meng2023distillation,
      author    = {Meng, Chenlin and Rombach, Robin and Gao, Ruiqi and Kingma, Diederik and Ermon, Stefano and Ho, Jonathan and Salimans, Tim},
    title     = {On Distillation of Guided Diffusion Models},
    booktitle = {Proceedings of the IEEE/CVF Conference on Computer Vision and Pattern Recognition (CVPR)},
    month     = {June},
    year      = {2023},
    pages     = {14297-14306}
}

@InProceedings{radford2021learning,
  title = 	 {Learning Transferable Visual Models From Natural Language Supervision},
  author =       {Radford, Alec and Kim, Jong Wook and Hallacy, Chris and Ramesh, Aditya and Goh, Gabriel and Agarwal, Sandhini and Sastry, Girish and Askell, Amanda and Mishkin, Pamela and Clark, Jack and Krueger, Gretchen and Sutskever, Ilya},
  booktitle = 	 {Proceedings of the 38th International Conference on Machine Learning},
  pages = 	 {8748--8763},
  year = 	 {2021},
  editor = 	 {Meila, Marina and Zhang, Tong},
  volume = 	 {139},
  series = 	 {Proceedings of Machine Learning Research},
  month = 	 {18--24 Jul},
  publisher =    {PMLR},
  url = 	 {https://proceedings.mlr.press/v139/radford21a.html}
}

@inproceedings{
dosovitskiy2021image,
title={An Image is Worth 16x16 Words: Transformers for Image Recognition at Scale},
author={Alexey Dosovitskiy and Lucas Beyer and Alexander Kolesnikov and Dirk Weissenborn and Xiaohua Zhai and Thomas Unterthiner and Mostafa Dehghani and Matthias Minderer and Georg Heigold and Sylvain Gelly and Jakob Uszkoreit and Neil Houlsby},
booktitle={International Conference on Learning Representations},
year={2021},
url={https://openreview.net/forum?id=YicbFdNTTy}
}

@InProceedings{sohldickstein2015deep,
  title = 	 {Deep Unsupervised Learning using Nonequilibrium Thermodynamics},
  author = 	 {Sohl-Dickstein, Jascha and Weiss, Eric and Maheswaranathan, Niru and Ganguli, Surya},
  booktitle = 	 {Proceedings of the 32nd International Conference on Machine Learning},
  pages = 	 {2256--2265},
  year = 	 {2015},
  editor = 	 {Bach, Francis and Blei, David},
  volume = 	 {37},
  series = 	 {Proceedings of Machine Learning Research},
  address = 	 {Lille, France},
  month = 	 {07--09 Jul},
  publisher =    {PMLR},
  url = 	 {https://proceedings.mlr.press/v37/sohl-dickstein15.html}
}

@article{cifar10,
title= {CIFAR-10 (Canadian Institute for Advanced Research)},
journal= {},
author= {Alex Krizhevsky and Vinod Nair and Geoffrey Hinton},
year= {},
url= {http://www.cs.toronto.edu/~kriz/cifar.html},
terms= {}
}

@inproceedings{heusel2018gans,
 author = {Heusel, Martin and Ramsauer, Hubert and Unterthiner, Thomas and Nessler, Bernhard and Hochreiter, Sepp},
 booktitle = {Advances in Neural Information Processing Systems},
 editor = {I. Guyon and U. Von Luxburg and S. Bengio and H. Wallach and R. Fergus and S. Vishwanathan and R. Garnett},
 pages = {},
 publisher = {Curran Associates, Inc.},
 title = {GANs Trained by a Two Time-Scale Update Rule Converge to a Local Nash Equilibrium},
 url = {https://proceedings.neurips.cc/paper_files/paper/2017/file/8a1d694707eb0fefe65871369074926d-Paper.pdf},
 volume = {30},
 year = {2017}
}

@InProceedings{pmlr-v139-radford21a,
  title = 	 {Learning Transferable Visual Models From Natural Language Supervision},
  author =       {Radford, Alec and Kim, Jong Wook and Hallacy, Chris and Ramesh, Aditya and Goh, Gabriel and Agarwal, Sandhini and Sastry, Girish and Askell, Amanda and Mishkin, Pamela and Clark, Jack and Krueger, Gretchen and Sutskever, Ilya},
  booktitle = 	 {Proceedings of the 38th International Conference on Machine Learning},
  pages = 	 {8748--8763},
  year = 	 {2021},
  editor = 	 {Meila, Marina and Zhang, Tong},
  volume = 	 {139},
  series = 	 {Proceedings of Machine Learning Research},
  month = 	 {18--24 Jul},
  publisher =    {PMLR},
  url = 	 {https://proceedings.mlr.press/v139/radford21a.html}
}

@inproceedings{
schuhmann2022laionb,
title={{LAION}-5B: An open large-scale dataset for training next generation image-text models},
author={Christoph Schuhmann and Romain Beaumont and Richard Vencu and Cade W Gordon and Ross Wightman and Mehdi Cherti and Theo Coombes and Aarush Katta and Clayton Mullis and Mitchell Wortsman and Patrick Schramowski and Srivatsa R Kundurthy and Katherine Crowson and Ludwig Schmidt and Robert Kaczmarczyk and Jenia Jitsev},
booktitle={Thirty-sixth Conference on Neural Information Processing Systems Datasets and Benchmarks Track},
year={2022},
url={https://openreview.net/forum?id=M3Y74vmsMcY}
}

@article{larkin2016reflections,
  title={Reflections on shannon information: In search of a natural information-entropy for images},
  author={Larkin, Kieran G},
  journal={arXiv preprint arXiv:1609.01117},
  year={2016}
}

@article{wu2013local,
  title={Local Shannon entropy measure with statistical tests for image randomness},
  author={Wu, Yue and Zhou, Yicong and Saveriades, George and Agaian, Sos and Noonan, Joseph P and Natarajan, Premkumar},
  journal={Information Sciences},
  volume={222},
  pages={323--342},
  year={2013},
  publisher={Elsevier}
}

@inproceedings{vila2014analysis,
  title={Analysis of image informativeness measures},
  author={Vila, Marius and Bardera, Anton and Feixas, Miquel and Bekaert, Philippe and Sbert, Mateu},
  booktitle={2014 IEEE International Conference on Image Processing (ICIP)},
  pages={1086--1090},
  year={2014},
  organization={IEEE}
}
